\documentclass[sigconf]{acmart}
\usepackage{microtype}
\usepackage{graphicx}
\usepackage{amsmath}
\usepackage{multirow}
\usepackage{booktabs}
\usepackage{algorithmic}
\usepackage[ruled]{algorithm2e}
\AtBeginDocument{%
  \providecommand\BibTeX{{%
    \normalfont B\kern-0.5em{\scshape i\kern-0.25em b}\kern-0.8em\TeX}}}

\begin{document}

\title{Coherence-Based Distributed Document Representation Learning for Scientific Documents}

\newcommand{\tsc}[1]{\textsuperscript{#1}}

\author{Shicheng Tan}
\author{Shu Zhao}\authornote{Corresponding authors}
\author{Yanping Zhang}\authornotemark[1]
\affiliation{
  \institution{Key Laboratory of Intelligent Computing and Signal Processing, Ministry of Education, Anhui Province 230601, P.R. China}
  \institution{School of Computer Science and Technology, Anhui University, Hefei, Anhui Province 230601, P.R. China}
  \institution{Information Materials and Intelligent Sensing Laboratory of Anhui Province, Anhui Province 230601, P.R. China}
  \country{}
}
\email{tsctan@foxmail.com,zhaoshuzs2002@hotmail.com,zhangyp2@gmail.com}

\renewcommand{\shortauthors}{Shicheng Tan}

\begin{abstract}
  Distributed document representation is one of the basic problems in natural language processing. Currently distributed document representation methods mainly consider the context information of words or sentences. These methods do not take into account the coherence of the document as a whole, e.g., a relation between the paper title and abstract, headline and description, or adjacent bodies in the document. The coherence shows whether a document is meaningful, both logically and syntactically, especially in scientific documents (papers or patents, etc.). In this paper, we propose a coupled text pair embedding (CTPE) model to learn the representation of scientific documents, which maintains the coherence of the document with coupled text pairs formed by segmenting the document. First, we divide the document into two parts (e.g., title and abstract, etc) which construct a coupled text pair. Then, we adopt negative sampling to construct uncoupled text pairs whose two parts are from different documents. Finally, we train the model to judge whether the text pair is coupled or uncoupled and use the obtained embedding of coupled text pairs as the embedding of documents. We perform experiments on three datasets for one information retrieval task and two recommendation tasks. The experimental results verify the effectiveness of the proposed CTPE model.
\end{abstract}


\begin{CCSXML}
<ccs2012>
   <concept>
       <concept_id>10010405.10010497.10010498</concept_id>
       <concept_desc>Applied computing~Document searching</concept_desc>
       <concept_significance>500</concept_significance>
       </concept>
   <concept>
       <concept_id>10010147.10010178.10010179</concept_id>
       <concept_desc>Computing methodologies~Natural language processing</concept_desc>
       <concept_significance>500</concept_significance>
       </concept>
   <concept>
       <concept_id>10010147.10010178.10010179.10003352</concept_id>
       <concept_desc>Computing methodologies~Information extraction</concept_desc>
       <concept_significance>300</concept_significance>
       </concept>
   <concept>
       <concept_id>10010405.10010497.10010498</concept_id>
       <concept_desc>Applied computing~Document searching</concept_desc>
       <concept_significance>300</concept_significance>
       </concept>
   <concept>
       <concept_id>10002951.10003317.10003318</concept_id>
       <concept_desc>Information systems~Document representation</concept_desc>
       <concept_significance>500</concept_significance>
       </concept>
 </ccs2012>
\end{CCSXML}

\ccsdesc[500]{Information systems~Document representation}
\ccsdesc[500]{Computing methodologies~Natural language processing}
\ccsdesc[300]{Applied computing~Document searching}
\ccsdesc[300]{Computing methodologies~Information extraction}

\keywords{document representation learning, coherence, neural networks, information retrieval, scientific documents}

\settopmatter{printacmref=false}  % 删除开头的 ACM Reference Format:
%\setcopyright{none}  % 删除版权声明
\renewcommand\footnotetextcopyrightpermission[1]{}
\pagestyle{plain}  % 参考 https://tex.stackexchange.com/questions/346292/how-to-remove-conference-information-from-the-acm-2017-sigconf-template

\maketitle

\section{Introduction}
Distributed document representation aims some basic problems such as retrieval, classification, and inference to support other downstream tasks. Due to the semantic complexity of the documentation, learning document representation is still a difficult task, not well solved at present. For a document composed of multiple bodies, there should be a logical relationship and smoothness between its adjacent bodies. Therefore, there is a relation between the adjacent bodies in the document, which shows whether a document is meaningful, both logically and syntactically \cite{li2014model}. This relation is very important to the scientific document, without this relation the document will become meaningless. It is very necessary to maintain this relation in distributed document representation. This relation is called coherence.

Current distributed document representation methods are based on word embedding or sentence embedding. \citeauthor{yurochkin2019hierarchical} \shortcite{yurochkin2019hierarchical} treat document representation as a hierarchical optimal transport based on the word mover’s distance \cite{kusner2015word}, word embedding, and topic. \citeauthor{chen2019self} \shortcite{chen2019self} build the document embedding by averaging the sentence embedding and training through discriminator that determines whether a sentence belongs to a document. However, these methods based on the context information of words or sentences do not take into account the coherence of the document as a whole. The embedding of words and sentences can easily maintain coherence because there is the context (adjacent words or sentences) around the words and sentences. The context of the document exists inside the document, not around it. Maintaining the coherence of the document as a whole is a challenge.

In this paper, we propose a coupled text pair embedding (CTPE) model to learn distributed document representation by maintaining the coherence of the scientific document. Specifically, we first divide the scientific document into two parts (such as the title and abstract) to form a coupled text pair. We use some common methods to obtain word or sentence embedding for all coupled text pairs (such as word embedding for short documents and sentence embedding for long documents). Then, we adopt negative sampling to construct uncoupled text pairs whose two parts are randomly derived from different documents. Our model maintains the coherence of the document by training to determine whether the text pair is coupled or uncoupled. We input the word embedding or sentence embedding of the text pair into the model and finally obtain two vector representations of the coupled text pair as the document representation. This paper further proposes a method for calculating the similarity of the coupled text pairs to calculate the similarity between documents represented by the paired embeddings.

The main contributions of this paper are summarized as follows:

\begin{itemize}
	\item We introduce coherence into representation of scientific documents for the first time, which makes a document meaningful both logically and syntactically, based on two adjacent bodies in the document.
	\item We propose a novel form of embedding and a general model, called coupled text pair embedding (CTPE), to maintain document coherence and represent documents.
	\item We propose a document similarity calculation method for computing documents represented by coupled text pair embedding.
	\item We perform experiments through information retrieval and recommendation to show that our model is competitive to the state-of-the-art methods based on context information of words or sentences.
\end{itemize} 

\section{Related Work}
Current distributed document representation learning methods are not specifically used for scientific documents. In the following we describe the main distributed document representation learning methods.

\subsection{Word-level methods}
The word-level method uses words as calculation objects to represent the document. Early distributed document representation methods are mainly algebraic or probabilistic models such as TF-IDF \cite{salton1988term} and topic model \cite{blei2003latent,yang2018topic}. These models treat the document as bag of words which neglect other useful information such as context information of words. With the development of neural networks, it is easier to obtain context information about words, such as word2vec \cite{mikolov2013distributed}, lifelong domain word embedding \cite{xu2018lifelong} and XLNet \cite{yang2019xlnet}, etc. On this basis, \citeauthor{le2014distributed} \shortcite{le2014distributed} and \citeauthor{luo2019learning} \shortcite{luo2019learning} predict a word embedding through other word embedding and document embedding while learning learn document embedding. \citeauthor{arora2016simple} \shortcite{arora2016simple}, \citeauthor{chen2017efficient} \shortcite{chen2017efficient} and \citeauthor{schmidt2019improving} \shortcite{schmidt2019improving} treat the average word embedding as document embedding. \citeauthor{hansen2019contextually} \shortcite{hansen2019contextually} weight word embedding to obtain document embedding. \citeauthor{kusner2015word} \shortcite{kusner2015word} and \citeauthor{wu2018word} \shortcite{wu2018word} get the distance between different documents by calculating word mover's distance.

There are some methods that base on labels of specific tasks. \citeauthor{xiao2019label} \shortcite{xiao2019label} propose a document representation model base on label semantic information for multi-label text classification. \citeauthor{RN1484} \shortcite{RN1484} use citations between documents to train the transformer model and obtain representations of scientific documents.

\subsection{Sentence-level methods}
The sentence-level method uses sentences as calculation objects to represent the document. In addition to word embedding, sentence embedding \cite{hill2016learning,logeswaran2018efficient} can also be obtained from the context information of the sentence. The average of a sentence embedding or sentence embedding that treats a document as a sentence can be used to represent document embedding. \citeauthor{li2014model} \shortcite{li2014model} maintain coherence in sentence embedding based on sentence order, but does not form document embedding. \citeauthor{chen2019self} \shortcite{chen2019self} build the document embedding by averaging the sentence embedding and training through discriminator that determines whether a sentence belongs to a document.

These methods based on the context information of words or sentences do not take into account the coherence of the document as a whole.

\section{Problem Statement and Model Architecture}
In this section, we formulate the problem of distributed document representation and introduce the structure and technical details of coupled text pair embedding (CTPE) model for distributed document representation.

\begin{figure*}[h]
	\centering
	\includegraphics[width=1\textwidth]{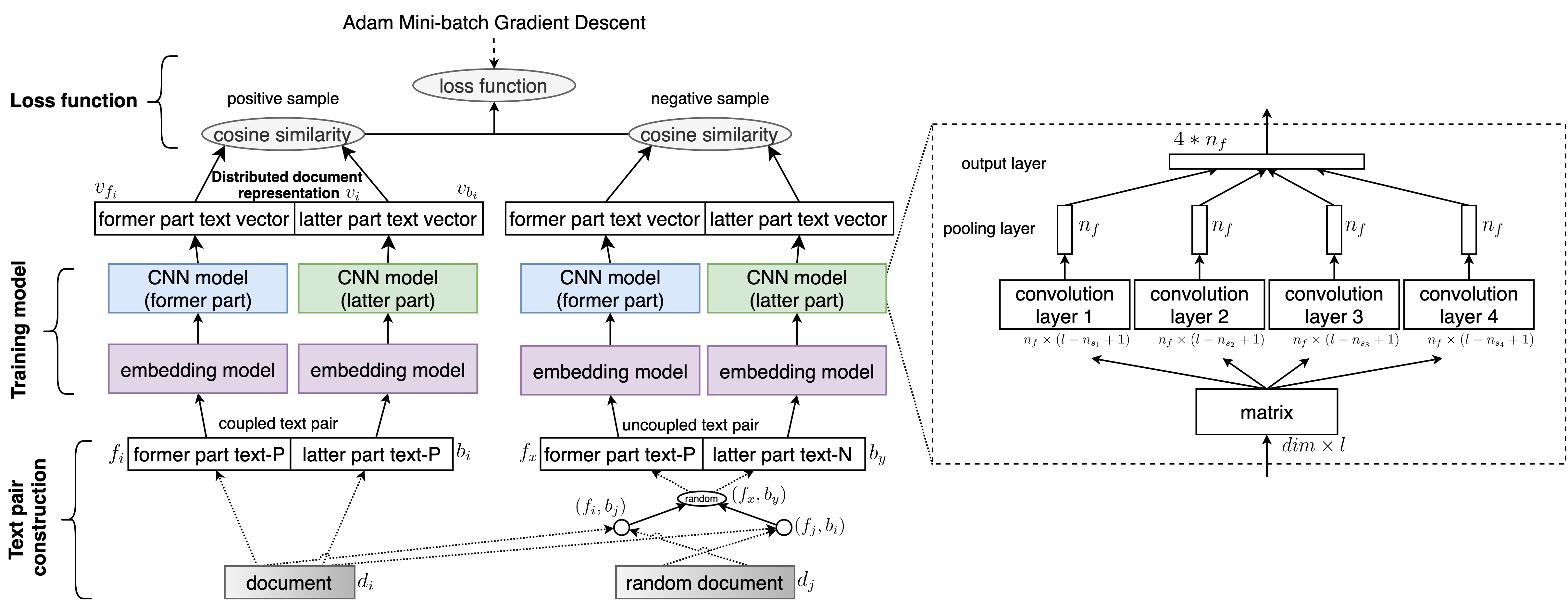}
	\caption{The overall architecture of CTPE. The positive sample takes $f_i$ and $b_i$ as example, and the negative sample takes $f_x$ and $b_y$ as example. $f_i$, $b_i$, $f_x$ and $b_y$ correspond to four CNN models (former part, latter part, former part and latter part) respectively, and then each CNN model has four parallel convolutional layers.}
	\label{model}
\end{figure*}

\subsection{Problem Statement}
Firse, we define our term in a formal way. For a documents set $\mathbf{D}=\{d_1,d_2,...,d_{|\mathbf{D}|}\}$, its coupled text pairs set is defined as $\mathbf{S}=\{s(d_i)|s(d_i)=(f_i,b_i),d_i\in\mathbf{D},i\in[1,|\mathbf{D}|]\}$ and its uncoupled text pairs set is defined as $\mathbf{S}^*=\{s_{ij}|s_{ij}=(f_i,b_j),1\leqslant i\neq j\leqslant|\mathbf{D}|\}$, where $f_i$ and $b_i$ denote the former and latter part of the text of document $d_i$, respectively. Document $d_i$ is the text we want to embed. According to the data available to us, documents can be the full text, the title and the abstract only, or the headline and bodies, etc. In order to get $(f_i,b_j)$ based on $d_i$, we define a parameter $pos$ for the segmentation position, which determines the split point that divides the document into two bodies. The left side of $pos$ is $f_i$, and the right side is $b_i$. For example, $pos$ is generally defined as the position between the title and abstract if the document only includes the title and the abstract.

We define our problem in a formal way. Given a documents set $\mathbf{D}$ and its coupled text pairs set $\mathbf{S}$. Our goal is to learn distributed representation $g[s(d_i)]=v_i=(v_{f_i},v_{b_i})$ for document $d_i$, where $v_{f_i}$ and $v_{b_i}$ denote embedding of $f_i$ and $b_i$, respectively.

In the next subsection, we will describe the proposed coupled text pair embedding (CTPE) model for distributed document representation.

\subsection{CTPE architecture}
Figure \ref{model} shows the overall architecture of CTPE. We introduce coherence into the document representation with the three parts of \textbf{text pair construction}, \textbf{training model}, and \textbf{loss function}, and propose a novel embedding form with the fourth part \textbf{distributed document representation}.

\textbf{Text pair construction.}
We propose a representation of text pairs so that we model learns to determine which document is more consistent because coherence is the relation between the two bodies in the document. We first divide each document $d_i$ into former part text $f_i$ and latter part text $b_i$ to form a coupled text pair. The $f_i$ and $b_i$ of the same document are used as a positive sample $(f_i,b_i)\in\mathbf{S}$. Then we randomly select the document $d_j=random(\mathbf{D})$ from the documents set $\mathbf{D}$ by negative sampling. We construct uncoupled text pairs $(f_i,b_j)$ and $(f_j,b_i)$ through $s(d_i)$ and $s(d_j)$. We randomly choose $(f_i,b_j)$ or $(f_j,b_i)$ as a negative sample $(f_x,b_y)=random[(f_i,b_j),(f_j,b_i)]\in\mathbf{S}^*$ whose $f_x$ and $b_y$ come from different documents. We adopt negative sampling to obtain coupled and uncoupled text pairs sets $\mathbf{S}$ and $\mathbf{S}^*$ in each epoch of training.

\textbf{Training model.}
We introduce a training model that learn the relation between the $f$ and $b$ to determine which document is more consistent. The embedding model aims to represent text pairs in a distributed form, such as word embedding, token embedding, sentence embedding, paragraph embedding model, or so on. We can choose different embedding models for documents of different lengths. We compare different embedding models in experiments. The embedding model converts text consisting of $l$ words, tokens, sentences, or paragraphs into $l$ vectors of dimension $dim$. The $dim*l$ dimension matrix after the embedding model is input to the CNN model to learn the relation between $f$ and $b$. The structure of the CNN model is composed of input layer, convolution layer, pooling layer, output layer. The number of channels of the CNN model is $n_f$, and the sizes of the convolution kernels are $n_s=\{n_{s1}, n_{s2},n_{s3},n_{s4}\}$. The right side of Figure \ref{model} shows the details of the CNN model. The implementation process training model is as follows.

Since the length of $f_i$ and $b_i$ are different, the maximum word length of the $f_i$ and $b_i$ is limited to $l_{max}\geqslant l$. First, for any $f_i$ and $b_i$ with an input length of $len\leqslant l_{max}$, a matrix composed of $l$ vector of $dim$ dimension is obtained from the embedding model, which is input to the input layer of the CNN model. Then, through the convolution operation of the rectified linear unit (ReLU) and the convolution filter that stride is 1, four convolutional layers can be obtained, and their sizes are respectively $n_f*(l-n_{s1}+1) $, $n_f*(l-n_{s2}+1)$, $n_f*(l-n_{s3}+1)$ and $n_f*(l-n_{s4}+1)$. Next, the convolutional layer is processed into a one-dimensional vector of length $n_f$ by the max pool operation. Finally, a text vector of length $4*n_f$ is obtained by cascading operation. The $(f_i, b_i)$ and the $(f_x, b_y)$ form vector pairs $(v_{f_i}, v_{b_i})$ and $(v_{f_x}, v_{b_y})$ with the embedding model and CNN model. In order to reduce the training time and space complexity, positive and negative samples usually share the same model parameters.

\textbf{Loss function.}
In order to maintain coherence in the document representation, the loss function aims to make the distance between the positive and negative samples larger, that is, the similarity between the $f_i$ and $b_i$ of the same document must be greater than the similarity between the $f_x$ and $b_y$ of the different documents. We assume a minimum similarity distance $M$, the difference in similarity between the positive and negative samples must be at least greater than $M$. If the difference of similarity between the positive and negative samples is greater than $M$, the similarity is recorded as the loss value, otherwise is not included in the loss value. For positive sample $(f_i, b_i)$, negative sample $(f_x,b_y)$, the loss function $L$ is defined as:
\begin{equation}
	\label{loss-function}
	\begin{aligned}
		L = &max\{0,-(\mathit{diff}_{xy}-M)\}\\
		where&~\mathit{diff}_{xy}=\\
		&\begin{cases}
			cos(v_{f_i},v_{b_i})-cos(v_{f_i},v_{b_j}),\; y=j\\
			cos(v_{f_i},v_{b_i})-cos(v_{f_j},v_{b_i}),\; x=j
		\end{cases}\\
	\end{aligned}
\end{equation}
where $cos$ denotes the cosine similarity of two vectors. $M$ represents the minimum similarity distance (called margin). This means that if $\mathit{diff}_{xy}$ is greater than $M$, then it is no longer necessary to modify the weight, i.e., $L=0$. The small batch gradient is reduced using the adaptive moment estimation (Adam) algorithm and the learning rate is set to $ln$.

\begin{figure}
	\centering
	\includegraphics[width=0.75\columnwidth]{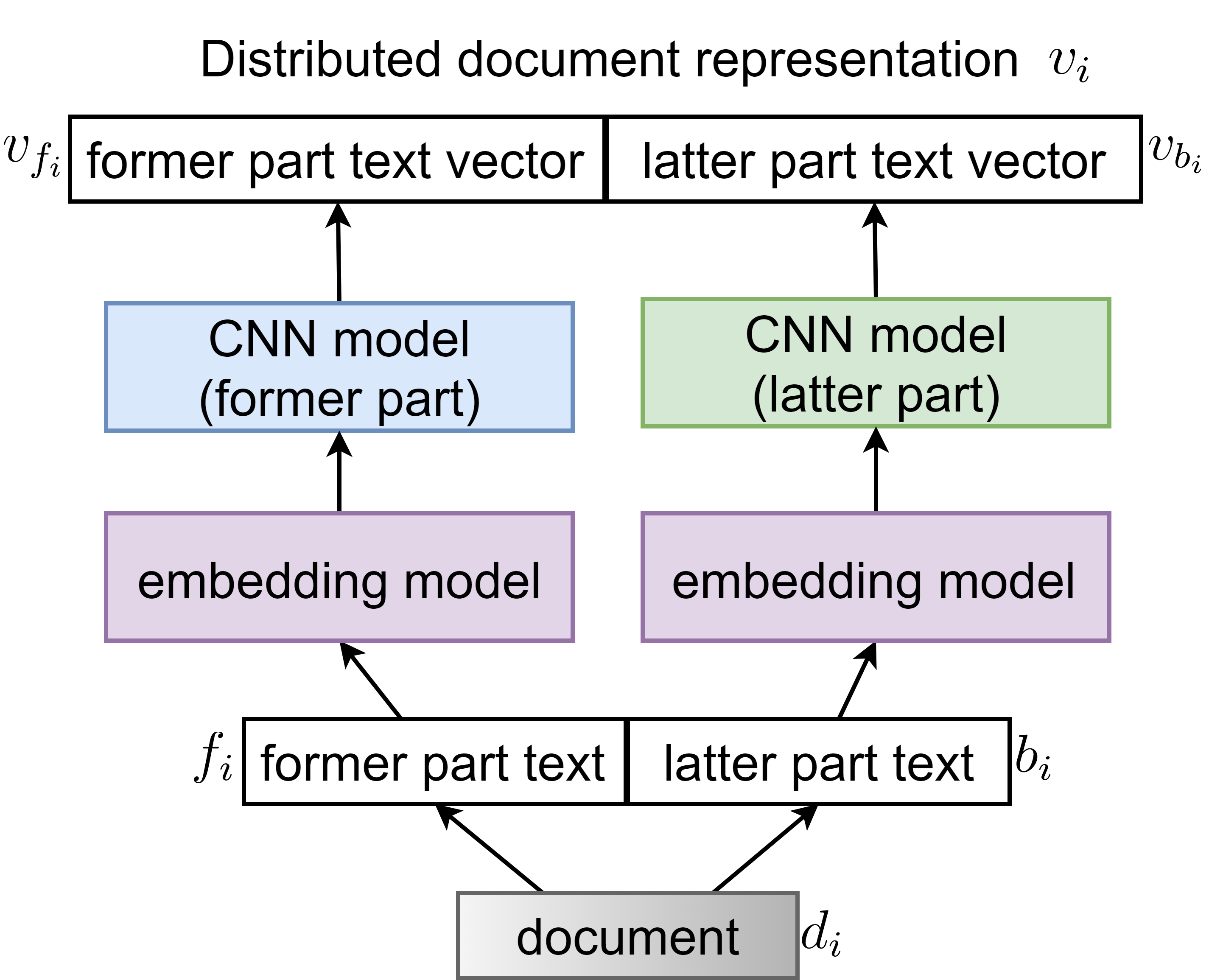}
	 \caption{Distributed document representation of CTPE.}
	\label{representation}
\end{figure}

\textbf{Distributed document representation.}
After training with the loss function, we need to obtain a distributed representation of the document with the model. As shown in Figure \ref{representation}, we calculate the vector pair $(v_{f_i},v_{b_i})$ of a document $d_i$ through the trained model of the previous section. $v_i=(v_{f_i},v_{b_i})$ is defined as the distributed representation of document $d_i$. $v_i$ is composed of a pair of vectors, which is different from a single embedded vector. The overall procedure from document to distributed representation is summarized in Algorithm \ref{CTPE}.

\begin{algorithm}
	\caption{\label{CTPE} Coupled text pair embedding (CTPE)}
	\LinesNumbered
	\KwIn{$d_k$, $\mathbf{D}$, $l_{max}$, $ln$, $L$, $pos$, embedding model $E$, batch size $b$, epoch $e$}
	\KwOut{document representation $v_k$}

	$\verb|//|$ Coupled text pair \\
	Initialize coupled text pair's set $\mathbf{P}$ \\
	\For{each $d_i\in\mathbf D$}
	{
		Obtain $f_i$ and $b_i$ based on $pos$ and $l_{max}$\\
		Coupled text pair $p(d_i)=(f_i,b_i)$\\
		$\mathbf{P}\leftarrow p(d_i)$
	}
	$\verb|//|$ Training\\
	\For{$i$th epoch, $i<e$}
	{
		Initialize uncoupled text pair's set $\mathbf{P^*}$ \\
		\For{each $d_i\in\mathbf D$}
		{
			$d_j=random(\mathbf D)$\\
			$(f_i,b_i),(f_j,b_j)=p(d_i),p(d_j)$\\
			$\mathbf{P^*}\leftarrow random[(f_i,b_j),(f_j,b_i)]$
		}
		\For{Select $b$ $data$ from $\mathbf{P},\mathbf{P^*}$} 
		{
			$\mathbf M\leftarrow 4\times b$ matrices$\leftarrow$ input $data$ into $E$\\
			$4\times b$ vectors $\leftarrow$ input $\mathbf M$ into CNN\\
			Adam processing by $L$ and $ln$
		}
	}
	$\verb|//|$ Distributed document representation \\
	$(f_k,b_k)=p(d_k)$\\
	$\mathbf M\leftarrow 2\times b$ matrices$\leftarrow$ enter $(f_k,b_k)$ into $E$\\
	$(v_{f_k},v_{b_k})\leftarrow 2\times b$ matrices$\leftarrow$ input $\mathbf M$ into CNN model\\
	\textbf{Return}: $v_k=(v_{f_k},v_{b_k})$
\end{algorithm}

\subsection{Calculation Method of Similarity}
\begin{figure}[!h]
	\centering
	\includegraphics[width=1\columnwidth]{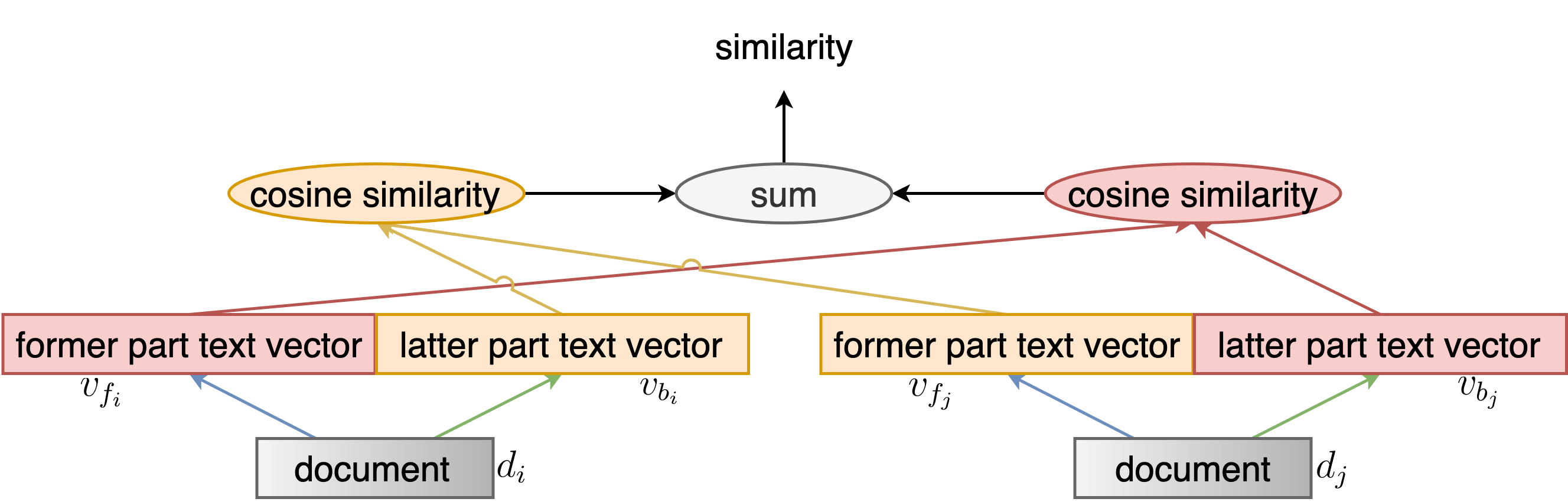}
	 \caption{Similarity calculation of documents in CTPE.}
	\label{similarity}
\end{figure}

We propose a novel similarity calculation method because of the special document representation. The general similarity calculation method only calculates the similarity without coherence. The calculation of similarity occurs between $f$ in one document and $b$ in another document, because coherence is the relation between the adjacent bodies. As shown in Figure \ref{similarity}, for any two documents $d_i$ and $d_j$, the similarity $sim(d_i,d_j)$ between them is defined as:
\begin{equation}
	\label{sim-function}
	\begin{aligned}
		&sim(d_i,d_j)=cos(v_{f_i},v_{b_j})+cos(v_{f_j},v_{b_i})
	\end{aligned}
\end{equation}
$sim(\cdot,\cdot)$ means the coherence between different bodies that come from different documents relative to the same document. $sim(\cdot,\cdot)$ is different from the general similarity and may not satisfy $sim(d_i,d_i)=sim(d_j,d_j)=2$ because it does not measure whether two documents are "exactly the same".

Finally, we obtain a distributed representation of the document, and a similarity calculation method for this distributed representation. Through these two operations, we implement downstream tasks.

\section{Experiments}
In this section, we evaluate the performance and effectiveness of our model. We conduct experiments compared with 24 comparison methods on arXiv (papers), DBLP (papers) and USPTO (patents) datasets. Our code will be released on GitHub\footnote{https://github.com/aitsc/text-representation}.

\begin{table*}
	\centering
	\begin{tabular}{ccccccc} \hline
	\multirow{2}{*}{}&\multicolumn{2}{c}{arXiv}&\multicolumn{2}{c}{DBLP}&\multicolumn{2}{c}{USPTO}\\
	\cmidrule(lr){2-3} \cmidrule(lr){4-5}\cmidrule(lr){6-7}
	  & candidate & test & candidate & test & candidate & test \\ \hline
	 documents & 20,000 & 1,000 & 20,000 & 1,000 & 20,000 & 1,000 \\
	 years & 1991-2015 & 2016 & 1967-2016 & 2017 & 2002-2016 & 2017-2018 \\
	 words per document & 205-425 & 205-418 & 155-421 & 156-412 & 250-680 & 255-629 \\ 
	 labels & - & 22-95 & - & 15-16 & - & 8-29 \\ \hline
	\end{tabular}
	\caption{\label{datasets} Datasets. The labels in the table refer to the range of the number of groundtruth candidate documents for the test document.}
\end{table*}

\subsection{Downstream Tasks and Datasets}
We use the arXiv dataset for the information retrieval task that aims to retrieve similar candidate papers for a paper (consisting of title and abstract). We use the DBLP and USPTO datasets for recommendation tasks (citation recommendation and patent recommendation). Citation recommendation task aims to recommend appropriate candidate citations for a paper (consisting of title and abstract). Patent recommendation task aims to recommend similar candidate patents for a patent (consisting of title, abstract, and claim). The details of datasets are described below.

\textbf{arXiv.} This dataset comes from the public data source of arxiv\footnote{ftp://3lib.org//oai\_dc/arxiv}, which contains a total of 1,180,081 papers. All papers contain titles, abstracts, authors, publication time, subject. For the test paper to be queried, candidate paper with the same subject as it is taken as groundtruth. Evaluation protocol: whether to retrieve papers that are candidates for groundtruth.

\textbf{DBLP.} This dataset comes from the public data source of  DBLP \cite{tang2008arnetminer}. We use the v10 version\footnote{https://aminer.org/citation}, which includes a total of 3,079,007 papers. We use titles, abstracts, publication times, and citations (groundtruth) in this dataset. Evaluation protocol: whether the groundtruth citation can be recommended.

\textbf{USPTO.} This dataset comes from the public data source of the PatentsView database\footnote{http://www.patentsview.org/download} (a total of 6,424,534 patents). We extracte the title, abstract, year, first paragraph of the claim, and the citation (groundtruth) cited by examiner from the dataset. Evaluation protocol: whether the groundtruth citation can be recommended.

In order to effectively evaluate the task, we perform experiments using cleaned subsets of arXiv, DBLP, and USPTO that are released on GitHub\footnote{https://github.com/opendata-ai/tr} in detail. Each dataset contains 20,000 candidate documents and 1000 test documents (papers or patents) randomly selected. Table \ref{datasets} describes these datasets in detail. The number of candidate and settings for DBLP and USPTO are similar to \cite{cai2018generative} and \cite{ji2019patent}.

In order to conviniently evaluate the performance of different comparison methods, we use the same text preprocessing method for all data. The text preprocessing method is divided into five steps: 1. Convert all text to lowercase; 2. Remove HTML labels; 3. Restore HTML escape characters; 4. Split text with punctuation; 5. Remove tokens without letters. For sentence embedding, we use the punctuation at the end of the sentence to segment the document into sentences before text preprocessing.

\begin{table*}[!h]
	\centering
	\begin{tabular}{cccccccccc} \hline
	\multirow{2}{*}{Methods}&\multicolumn{3}{c}{arXiv}&\multicolumn{3}{c}{DBLP}&\multicolumn{3}{c}{USPTO}\\
	\cmidrule(lr){2-4}\cmidrule(lr){5-7}\cmidrule(lr){8-10}
	  & P & R & F$_1$ & P & R & F$_1$ & P & R & F$_1$ \\ \hline
	 doc2vec(ran) & 0.0030 & 0.0009 & 0.0014 &  0.0007 & 0.0009 & 0.0008 &  0.0005 & 0.0010 & 0.0006 \\
	 avg-word2vec(ran) & 0.0134 & 0.0048 & 0.0071 & 0.0414 & 0.0542 & 0.0469 & 0.0457 & 0.0849 & 0.0594 \\ \hline
	 TF-IDF & 0.1532 & 0.0571 & 0.0831 & 0.2564 & 0.3360 & 0.2908 & 0.1280 & 0.2419 & 0.1674 \\
	 LSA & 0.0289 & 0.0098 & 0.0146 & 0.0420 & 0.0549 & 0.0476 & 0.0350 & 0.0653 & 0.0456 \\ 
	 LDA & 0.0361 & 0.0132 & 0.0193 & 0.0220 & 0.0287 & 0.0249 & 0.0309 & 0.0589 & 0.0406 \\ \hline
	 avg-GloVe & 0.1025 & 0.0369 & 0.0543 & 0.1564 & 0.2050 & 0.1774 & 0.1061 & 0.1996 & 0.1385 \\
	 avg-GloVe(full) & 0.1747 & 0.0634 & 0.0931 & 0.1821 & 0.2387 & 0.2066 & 0.1036 & 0.1959 & 0.1355 \\
	 avg-word2vec & 0.1609 & 0.0570 & 0.0841 & 0.2062 & 0.2702 & 0.2339 & 0.1376 & 0.2588 & 0.1796 \\
	 avg-word2vec(full) & 0.2551 & 0.0913 & 0.1345 & 0.2163 & 0.2834 & 0.2453& 0.1275 & 0.2419 & 0.1670 \\ 
	 WMD-GloVe & 0.1138 & 0.0413 & 0.0606 & 0.2081 & 0.2726 & 0.2360 & 0.1127 & 0.2120 & 0.1471 \\
	 WMD-GloVe(full) & 0.1393 & 0.0514 & 0.0751 & 0.2225 & 0.2915 & 0.2524 & 0.1185 & 0.2226 & 0.1547 \\ 
	 WMD-word2vec & 0.1637 & 0.0602 & 0.0880 & 0.2482 & 0.3253 & 0.2816 & 0.1356 & 0.2563 & 0.1774 \\
	 WMD-word2vec(full) & 0.2223 & 0.0827 & 0.1206 & 0.2527 & 0.3310 & 0.2866& 0.1376 & 0.2605 & 0.1801 \\ 
	 doc2vec & 0.1272 & 0.0462 & 0.0677 & 0.1554 & 0.2036 & 0.1763 & 0.1014 & 0.1914 & 0.1326 \\
	 Doc2VecC & 0.2825 & 0.1030 & 0.1510 & 0.2579 & 0.3380 & 0.2926 & 0.1635 & 0.3094 & 0.2140 \\ \hline
	 avg-Skip-thoughts & 0.0578 & 0.0231 & 0.0330 & 0.0699 & 0.0918 & 0.0794 & 0.0723 & 0.1364 & 0.0945 \\ \hline
	 CTPE-word2vec(full) & \bf{0.3066} & \bf{0.1124} & \bf{0.1645} & \bf{0.2808} & \bf{0.3680} & \bf{0.3185} & \bf{0.1889} & \bf{0.3609} & \bf{0.2480} \\ \hline
	\end{tabular}
	\caption{\label{P} Experimental measured as P, R and F$_1$.}
\end{table*}

\begin{table*}[!h]
	\centering
	\begin{tabular}{cccccccccc} \hline
	\multirow{2}{*}{Methods}&\multicolumn{3}{c}{arXiv}&\multicolumn{3}{c}{DBLP}&\multicolumn{3}{c}{USPTO}\\
	\cmidrule(lr){2-4}\cmidrule(lr){5-7}\cmidrule(lr){8-10}
	  & MAP & NDCG & bpref & MAP & NDCG & bpref & MAP & NDCG & bpref \\ \hline
	 doc2vec(ran) & 0.0007 & 0.0034 & 0.4767 &   0.0001 & 0.0007 & 0.4753 & 0.0001 & 0.0004 & 0.4753 \\
	 avg-word2vec(ran) & 0.0041 & 0.0162 & 0.4834 & 0.0221 & 0.0642 & 0.5051 & 0.0276 & 0.0671 & 0.5081 \\ \hline
	 TF-IDF & 0.0773 & 0.1751 & 0.5666 & 0.1692 & 0.3257 & 0.6475 & 0.0745 & 0.1681 & 0.5624 \\
	 LSA & 0.0073 & 0.0312 & 0.4912  & 0.0188 & 0.0579 & 0.5039 & 0.0189 & 0.0487 & 0.4994 \\ 
	 LDA & 0.0093 & 0.0367 & 0.4944 & 0.0072 & 0.0267 & 0.4890 & 0.0141 & 0.0401 & 0.4954 \\ \hline
	 avg-GloVe & 0.0425 & 0.1173 & 0.5370  & 0.0917 & 0.2083 & 0.5819 & 0.0590 & 0.1404 & 0.5480 \\
	 avg-GloVe(full) & 0.0875 & 0.1983 & 0.5786 & 0.1084 & 0.2368 & 0.5974 & 0.0596 & 0.1383 & 0.5464 \\
	 avg-word2vec &0.0740 & 0.1779 & 0.5686 & 0.1280 & 0.2641 & 0.6140 & 0.0796 & 0.1764 & 0.5672 \\
	 avg-word2vec(full) & 0.1492 & 0.2810 & 0.6241 & 0.1367 & 0.2786 & 0.6211 & 0.0710 & 0.1629 & 0.5607 \\ 
	 WMD-GloVe & 0.0527 & 0.1393 & 0.5466 & 0.1406 & 0.2837 & 0.6216 & 0.0686 & 0.1527 & 0.5547 \\
	 WMD-GloVe(full) & 0.0660 & 0.1659 & 0.5603 & 0.1510 & 0.2993 & 0.6308 &  0.0719 & 0.1596 & 0.5577 \\
	 WMD-word2vec & 0.0797 & 0.1880 & 0.5736 & 0.1714 & 0.3280 & 0.6474 & 0.0826 & 0.1792 & 0.5693 \\
	 WMD-word2vec(full) & 0.1291 & 0.2556 & 0.6095 & 0.1733 & 0.3311 & 0.6490 & 0.0837 & 0.1822 & 0.5707 \\ 
	 doc2vec & 0.0514 & 0.1420 & 0.5491 & 0.0857 & 0.2027 & 0.5790 & 0.0566 & 0.1350 & 0.5444 \\ 
	 Doc2VecC & 0.1727 & 0.3056 & 0.6358 & 0.1719 & 0.3239 & 0.6475 &  0.0964 & 0.2062 & 0.5841 \\ \hline
	 avg-Skip-thoughts & 0.0230 & 0.0676 & 0.5104 & 0.0387 & 0.1044 & 0.5254 & 0.0407 & 0.1005 & 0.5265 \\ \hline
	 CTPE-word2vec(full) & \bf{0.1893} & \bf{0.3233} & \bf{0.6474}  & \bf{0.1846} & \bf{0.3476} & \bf{0.6595} & \bf{0.1146} & \bf{0.2381} & \bf{0.6020} \\ \hline
	\end{tabular}
	\caption{\label{MAP} Experimental measured as MAP, NDCG and bpref.}
\end{table*}

\subsection{Comparison Methods}
We compare our model with the following three types of unsupervised distributed document representation methods on retrieval and recommendation tasks.

\textbf{Random embedding model.}

We generate random embedding (based on uniform distribution) of documents and words as baseline methods, respectively called doc2vec(ran) and word2vec(ran). The avg-word2vec(ran) use average word embeddings as a document embedding. We use randomly generated embedding to show the difficulty of the task.

\textbf{Word-level methods.}

Algebraic or probabilistic model: TF-IDF \cite{salton1988term}, LSA \cite{deerwester1990indexing} and LDA \cite{blei2003latent}.

Word embedding based model: GloVe \cite{pennington2014glove}, word2vec \cite{mikolov2013distributed}, doc2vec \cite{le2014distributed}, WMD \cite{kusner2015word} and Doc2VecC \cite{chen2017efficient}. Since the corpus of datasets is a subset of the public data source, we train word2vec and GloVe on the experimental dataset and the public data source, called avg-word2vec and avg-word2vec(full) and avg-GloVe and avg-GloVe(full).

Deep language model: ELMo \cite{peters2018deep}, GPT \cite{radford2018improving}, GPT-2 \cite{radford2019language}, BERT \cite{devlin2019bert}, TransXL \cite{dai2019transformer}, XLM \cite{lample2019cross}, XLNet \cite{yang2019xlnet} and RoBERTa \cite{liu2019roberta}. We adopt pre-trained models without fine-tuning because of unsupervised document representation learning on these methods to verify the effectiveness of CTPE on different embedding models.

\textbf{Sentence-level methods.}

Sentence embedding based model: Skip-thoughts \cite{kiros2015skip}. The average of the sentence embeddings in the document is represented as the document embedding, called avg-Skip-thoughts.

\subsection{Metrics and Setup}
We use the methods to find the $topN$=20 results for three datasets and compare the result of each method with the labels in the datasets. We use the precision (P), recall (R), and F1 score (F$_1$) as evaluation metrics. We also employ several popular information retrieval measures \citep{buttcher2016information} including mean averaged precision (MAP), normalized discounted cumulative gain (NDCG), and bpref \citep{buckley2004retrieval}. These are popular measures for information retrieval \cite{zhang2020selective} and recommendation \cite{cai2018generative}. The parameters and setup of the comparison methods are described in detail below.

\textbf{random embedding.} This is a baseline method for randomly getting text embedding. We generate random embedding of documents and words, respectively called doc2vec(ran) and word2vec(ran). The embedded dimensions of both methods are set to 100.

\textbf{TF-IDF\footnote{https://scikit-learn.org/stable/modules/generated/ sklearn.feature\_extraction.\\text.TfidfVectorizer.html}.} The distributed representation of the document consists of the TF-IDF of each word in the document. The dimension of the document is the number of unique words in the dataset, and the cosine distance is use to calculate the similarity.

\textbf{LSA\footnote{https://radimrehurek.com/gensim/models/lsimodel.html}.} We set the number of topics and iterations to 100. Other parameters are consistent with gensim's default settings.

\textbf{LDA\footnote{https://radimrehurek.com/gensim/models/ldamodel.html}.} We set the number of topics and iterations to 100. Other parameters are consistent with gensim's default settings.

\textbf{word2vec\footnote{https://code.google.com/archive/p/word2vec}.} We use the average word embedding of word2vec as the document embedding, and the dimension is set to 100. The window size is set to 10, the maximum number of iterations is set to 100, and the strategy is set to skip-gram.

\textbf{GloVe\footnote{https://github.com/stanfordnlp/GloVe}.} We use the average word embedding of GloVe as the document embedding, and the dimension is set to 100. Since the experimental dataset is a subset of all the data, we train the word embedding on the experimental data set and all the data, called avg-GloVe and avg-GloVe(full). The window size is set to 10 and the maximum number of iterations is set to 100.

\textbf{doc2vec\footnote{https://radimrehurek.com/gensim/models/doc2vec.html}.} The dimension is set to 100. The window size is set to 10. The number of iterations is set to 50. The strategy is set to PV-DM.

\textbf{WMD\footnote{https://github.com/src-d/wmd-relax}.} The WMD is a distance measurement algorithm based on word embedding. We evaluate the WMD based on GloVe and word2vec, including WMD-word2vec, WMD-word2vec(full), WMD-GloVe, WMD-GloVe(full).

\textbf{Doc2VecC\footnote{https://github.com/mchen24/iclr2017}.} Doc2VecC is based on word2vec, where the settings for the word2vec part are consistent with the word2vec model above. The sentence sample is set to 0.1.

\textbf{Skip-thoughts\footnote{https://github.com/ryankiros/skip-thoughts\#getting-started}.} Skip-thoughts is a sentence embedding model. We use the generic model trained on a much bigger book corpus to encode the sentences. A vector of 4800 dimensions, first 2400 from the uni-skip model, and the last 2400 from the bi-skip model, are generated for each sentences. The EOS token is used. The average of the sentence embeddings in the document is represented as the embedding of the document.

\textbf{ELMo, GPT, GPT-2, BERT, Transformer XL, XLM, XLNet, RoBERTa.} The distributed document representation is represented by the average pooling of token embedding (BERT does not include [CLS] and [SEP]). The pre-trained models used in this paper are Original (5.5B)\footnote{https://allennlp.org/elmo}, openai-gpt\footnote{https://huggingface.co/pytorch-transformers/pretrained\_models.html\label{web}}, gpt2-medium\textsuperscript{\ref{web}}, bert-large-uncased-whole-word-masking\textsuperscript{\ref{web}}, transfo-xl-wt103\textsuperscript{\ref{web}}, xlm-mlm-en-2048\textsuperscript{\ref{web}}, xlnet-large-cased\textsuperscript{\ref{web}}, and roberta.large\footnote{https://github.com/pytorch/fairseq/tree/master/examples/\\roberta}, respectively. Because these deep language models are context-dependent token embedding, there are two strategies for training text pairs: the former part text and the latter part text are trained separately or together. We take the maximum of the performance of these two strategies.

We use Tensorflow to implement our model. We set batch size to 200. The parameters in CTPE include $l$, $l_{max}$, $n_s$, $n_f$, $ln$, $M$ and $dim$, which are empirically set to 200, 200, 1024,\{1,2,3,5\}, 0.001, 0.1, 100, respectively. If the embedding model is a sentence embedding model, $l$ is set to 20. The word embedding dimension of all models is set to 100, and other parameters are set to the default recommendation. All comparison methods use cosine similarity. In order to apply to various unsupervised tasks, there is no verification set and the parameters of different tasks are consistent.

Our method and all comparison methods use all documents in the dataset for unsupervised training and obtain all document embeddings. Test documents with groundtruth are used to test performance. Our model is trained on the RTX 2080TI. The training time for each result is limited to one day. There is no verification set here. If the loss function value does not decrease for more than 12 hours, the training will be stopped early.

\subsection{Experimental Results}
\label{sec:results}
The CTPE model uses word2vec(full) as the embedding model. We discuss other embedding models in Section \ref{sec:ablation}. For the arXiv and DBLP datasets, $pos$ is set between the title and abstract. For the USPTO dataset, $pos$ is set between abstract and claim. We discuss other $pos$ in section \ref{sec:segmentation}. Table \ref{P} and Table \ref{MAP} show the results of 17 methods on 3 datasets. From these results, we have the following observations and analysis.

CTPE has the best performance. On the arXiv dataset, our model outperforms Doc2VecC by 2.41\% and 1.66\% on precision and MAP. On the DBLP dataset, our model outperforms Doc2VecC by 2.29\% and 1.27\% on precision and MAP. On the USPTO dataset, our model outperforms Doc2VecC by 2.54\% and 1.82\% on precision and MAP. The CTPE model maintains the coherence of the document as a whole, so it achieves the best performance. Normal documents are coherent, otherwise they don't make sense.

Here are some other results and facts that need to be explained: (1) The performance of avg-word2vec(ran) is better than doc2vec(ran), because doc2vec(ran) is completely random, and avg-word2vec(ran) can uniquely identify a word. (2) The arXiv dataset has lower recall than other datasets. Because the arxiv dataset has more labels (see Table \ref{datasets}), it is more difficult to find all labels. (3) The word2vec-based approach is better than the GloVe-based approach on all three datasets. (4) The algorithm that uses all the data (full) is mostly better than the algorithm using the experimental dataset, which verifies the impact of data quality on the algorithm. (5) The WMD is better than average word embedding in most experiments, but not all, such as WMD-w(full) and WMD-g(full) on arXiv, because the WMD based on word embedding does not learn new information from the document.

\begin{table*}[!h]
	\centering
	\begin{tabular}{cccccccccc} \hline
	\multirow{2}{*}{Methods}&\multicolumn{3}{c}{arXiv}&\multicolumn{3}{c}{DBLP}&\multicolumn{3}{c}{USPTO}\\
	\cmidrule(lr){2-4}\cmidrule(lr){5-7}\cmidrule(lr){8-10}
	  & avg & CTPE$_0$ & CTPE & avg & CTPE$_0$ & CTPE & avg & CTPE$_0$ & CTPE \\ \hline
	 Skip-thoughts & 0.0330 & 0.0007 & \bf{0.0623} & 0.0794 & 0.0015 & \bf{0.1302} & 0.0945 & 0.0067 & \bf{0.1142} \\ \hline
	 word2vec(ran) & 0.0071 & 0.0034 & \bf{0.0637} & 0.0469 & 0.0022 & \bf{0.2277} & 0.0594 & 0.0549 & \bf{0.1378} \\
	 GloVe & 0.0543 & 0.0137 & \bf{0.0752} & 0.1774 & 0.0212 & \bf{0.2867} & 0.1385 & 0.1581 & \bf{0.2220} \\
	 GloVe(full) & 0.0931 & 0.0209 & \bf{0.1230} & 0.2066 & 0.0442 & \bf{0.2941} & 0.1355 & 0.1617 & \bf{0.2215} \\
	 word2vec & 0.0841 & 0.0555 & \bf{0.1098} & 0.2339 & 0.1390 & \bf{0.3002} & 0.1796 & 0.2182 & \bf{0.2381} \\
	 word2vec(full) & 0.1345 & 0.1093 & \bf{0.1645} & 0.2453 & 0.1607 & \bf{0.3185} & 0.1670 & 0.2095 & \bf{0.2480} \\ \hline
	 ELMo & 0.0915 & 0.0356 & \bf{0.1099} & 0.1346 & 0.0230 & \bf{0.2428} & 0.1349 & 0.1198 & \bf{0.1965} \\
	 GPT & 0.0676 & 0.0080 & \bf{0.0730} & 0.1910 & 0.0288 & \bf{0.1962} & 0.1651 & 0.0987 & \bf{0.1780} \\
	 GPT-2 & 0.0423 & 0.0011 & \bf{0.0738} & 0.0638 & 0.0007 & \bf{0.2027} & 0.0531 & 0.0021 & \bf{0.1537} \\
	 BERT & 0.1139 & 0.0267 & \bf{0.1280} & 0.1824 & 0.0095 & \bf{0.2639} & 0.1628 & 0.0991 & \bf{0.2130} \\
	 TransXL & 0.0190 & 0.0105 & \bf{0.0717} & 0.1489 & 0.0137 & \bf{0.1771} & 0.1626 & 0.1284 & \bf{0.1641} \\
	 XLM & 0.1226 & 0.0315 & \bf{0.1292} & 0.2107 & 0.0258 & \bf{0.2816} & 0.1753 & 0.1163 & \bf{0.2200} \\
	 XLNet & 0.0792 & 0.0060 & \bf{0.1083} & 0.2107 & 0.0121 & \bf{0.2412} & 0.1753 & 0.0097 & \bf{0.2137} \\
	 RoBERTa & 0.0616 & 0.0020 & \bf{0.0684} & 0.1493 & 0.0038 & \bf{0.2255} & 0.1190 & 0.0454 & \bf{0.1624} \\ \hline
	\end{tabular}
	\caption{\label{avg-CTPE} Experimental results based on different embedding models on F$_1$.}
\end{table*}

\subsection{Ablation Analysis}
\label{sec:ablation}
We conduct an ablation analysis on CTPE to examine the effectiveness of each component. We conduct experiments on three components: (1) We use different embedding models in the \textbf{training model}. (2) A model called CTPE$_0$ that does not use the \textbf{loss function} to train. (3) A \textbf{distributed document representation} method (avg) that uses average embedding instead of coupled text pair representation. Table \ref{avg-CTPE} shows the impact of 14 embedding models on the performance of avg, CTPE$_0$ and CTPE in 3 datasets using F$1$ metric. We have the following observations and analysis:

(1) CTPE improves the performance of various types of embedding models (word and sentence embedding models) effectively. CTPE outperforms avg model on average by 2.6\%, 7.9\%, and 5.4\% on arXiv, DBLP, and USPTO, because the avg model does not consider coherence. (2) The performance of CTPE is significantly better than CTPE$_0$, which proves the effectiveness of the loss function because the loss function guides the training model to determine which documents are more coherent. (3) Our distributed document representation cannot be directly used for general embedding models. More than 90\% of avg methods perform better than CTPE$_0$ models, because our similarity calculation method is dependent on coherence. (4) The performance of CTPE-Skip-thoughts is higher than that of avg-Skip-thoughts, which means that CTPE is suitable for super-long documents. CTPE can be based on embedding models of different granularity, such as sentence and paragraph embedding, which is several orders of magnitude less complex than CTPE that requires input of all words.

The performance of word embedding is stronger than the deep language model because the latter is evaluated by unsupervised document representation without fine-tuning.

\begin{table*}[!h]
	\centering
	\begin{tabular}{ccccccccccc} \hline
	\multirow{2}{*}{$pos$}&\multicolumn{3}{c}{arXiv}&\multicolumn{3}{c}{DBLP}&\multicolumn{3}{c}{USPTO}&\multirow{2}{*}{avg}\\
	\cmidrule(lr){2-4}\cmidrule(lr){5-7}\cmidrule(lr){8-10}
	  & P & R & F$_1$ & P & R & F$_1$ & P & R & F$_1$ \\ \hline
	  20\% & 0.3265 & 0.1211 & 0.1767 & \textbf{0.2925} & \textbf{0.3835} & \textbf{0.3319} & 0.1855 & 0.3545 & 0.2435 & \bf{0.2684} \\
	  40\% & \textbf{0.3283} & \textbf{0.1219} & \textbf{0.1778} & 0.2786 & 0.3651 & 0.3160 & 0.1821 & 0.3485 & 0.2392 & 0.2620\\
	  60\% & 0.3149 & 0.1155 & 0.1690 & 0.2652 & 0.3477 & 0.3009 & 0.1628 & 0.3103 & 0.2136 & 0.2444\\
	  80\% & 0.2934 & 0.1054 & 0.1551 & 0.2319 & 0.3040 & 0.2631 & 0.1526 & 0.2893 & 0.1998 & 0.2216\\
	 m & 0.3066 & 0.1124 & 0.1645 & 0.2808 & 0.3680 & 0.3185 & \textbf{0.1889} & \textbf{0.3609} & \textbf{0.2480} & 0.2610\\ \hline
	\end{tabular}
	\caption{\label{split} Experimental results of CTPE on different segmentation positions. The embedding model is based on word2vec(full). The above m and avg denote the meaningful segmentation and the average value of the row.}
\end{table*}

\begin{table*}[!h]
	\centering
	\begin{tabular}{cccccccccc} \hline
	\multirow{2}{*}{$pos$}&\multicolumn{3}{c}{arXiv}&\multicolumn{3}{c}{DBLP}&\multicolumn{3}{c}{USPTO}\\
	\cmidrule(lr){2-4}\cmidrule(lr){5-7}\cmidrule(lr){8-10}
	  & P & R & epochs & P & R & epochs & P & R & epochs \\ \hline
	  CTPE$_T$ & 0.2944 & 0.1074 & \textbf{3} & 0.2716 & 0.3561 & \textbf{78} & \textbf{0.1903} & \textbf{0.3628} & \textbf{9} \\
	  CTPE & \textbf{0.3066} & \textbf{0.1124} & 48 & \textbf{0.2808} & \textbf{0.3680} & 206 & 0.1889 & 0.3609 & 133 \\ \hline
	\end{tabular}
	\caption{\label{sampling} Experimental results of CTPE on different sampling methods. The embedding model is based on word2vec(full). Metrics include accuracy, recall, and training epochs of the result.}
\end{table*}

\subsection{Segmentation Position Analysis}
\label{sec:segmentation}
The setting of the parameter $pos$ in Section \ref{sec:results} is called the meaningful segmentation that divides the document into chapters or paragraphs, retaining complete paragraphs and sentences. We need to analyze how different segmentation positions affect the maintenance of coherence, as not all documents are suitable for this split method, such as some documents without meaningful segmentation. We examine the impact of different $pos$ on experimental performance. We divide the document into five equal parts and evaluate performance at four different $pos$ (20\%, 40\%, 60\%, 80\%). Our segmentation is at the token level. The meaningful segmentation positions (all samples average) of arXiv, DBLP, and USPTO are 4.7\%, 5.1\%, and 40.2\%, respectively. Table \ref{split} shows the experimental results of five positions on all datasets. We have the following observations and analysis.

 Experiments show that different segmentation positions have an impact on the maintaining of coherence because not all adjacent bodies in a document have strong coherence. Based on these results, we can draw three valuable suggestions: (1) If the document has meaningful segmentation, using meaningful segmentation can achieve excellent performance because meaningful segmentation outperforms all comparison methods, although not the best. (2) If the document does not have meaningful segmentation, using 20\%-40\% segmentation position can also achieve excellent performance because the performance of these $pos$ is the best. (3) Finding the best segmentation position will be a meaningful future work.

\subsection{Sampling with Imbalance Data}
In the process of text pair construction, negative samples are randomly combined from the former part text and latter part text of different documents, which makes the number of negative samples far more than positive samples. If the number of positive samples is $n$, then the number of negative samples is $n(n-1)$. Using negative sampling may ignore important negative samples and take longer.

We propose TF-IDF sampling to analyze the efficiency and performance of different sampling methods. In negative sampling, for document $d_i$, we choose document $d_j=random(\mathbf{D})$. In the TF-IDF sampling, for the document $d_i$, we select the document $d_j=random(random(T_{100}(d_i,\mathbf{D})),random(\mathbf{D}))$, where $T_{100}(d_i,\mathbf{D})$ denotes the top 100 documents in $\mathbf{D}$ that are most similar (TF-IDF similarity) to $d_i$. This sampling method is based on three considerations: (1) We introduce $T_{100}(d_i,\mathbf{D})$ to reduce the probability of ignoring important negative samples. (2) It is easier to overfit by training with only 100 documents instead of all documents. (3) The documents obtained with TF-IDF cannot completely represent the real similar documents. If $random(\mathbf{D})$ is not used, the model will overfit the results of TF-IDF.

Table \ref{sampling} shows the experimental results of CTPE and CTPE$_T$ (TF-IDF sampling) on all datasets. We draw a conclusion from these results: CTPE has better performance, but CTPE$_T$ is faster. The TF-IDF sampling allows the model to better distinguish the differences between similar documents, so it costs less epochs for training, but it is easier to overfit the samples sampled by TF-IDF. If we have higher requirements for training speed, CTPE$_T$ is more suitable, otherwise CTPE is more suitable.

\section{Conclusions}
In this paper, we propose a coupled text pair embedding (CTPE) model for distributed document representation, a novel architecture that is able to maintain the coherence of the scientific document for better document representing. We divide the scientific document into two parts and obtain a distributed representation of the document through CNN and embedding model. We use the coupling relationship between the two parts of the document to train the model. 

In the future, we will focus on the improvement of model structure and the application of supervised tasks (e.g., supervised information retrieval, text classification, or clickbait and fake news identification, etc). We will try more elaborate document encodings (e.g., perhaps splitting documents into more than two parts). There are also two interesting future works that are the precise measure of coherence and automatic selection of $pos$. For natural language processing tasks with more supervisory information, we will perform supervised fine-tuning on specific tasks based on the pre-training model of CTPE. In addition, CTPE model is a good choice for some tasks that lack supervisory information.

\begin{acks}
This work was partially supported by National High Technology Research and Development Program (Grant \# 2017YFB1401903), the National Natural Science Foundation of China (Grants \# 61876001, \# 61602003 and \# 61673020), the Provincial Natural Science Foundation of Anhui Province (Grants \# 1708085QF156), and the Recruitment Project of Anhui University for Academic and Technology Leader.
\end{acks}

\bibliographystyle{ACM-Reference-Format}
\bibliography{14}

\end{document}